\documentclass[letterpaper, 10pt, conference]{ieeeconf}  

\IEEEoverridecommandlockouts                              

\usepackage{color}
\usepackage{colortbl}
\usepackage{tabularx}
\usepackage{tcolorbox}
\usepackage{courier}
\usepackage{graphbox}
\usepackage{booktabs}
\usepackage{amsmath,amsfonts}
\usepackage{graphicx} 
\usepackage{subcaption}
\usepackage{tabularray}
\usepackage{caption}
\usepackage{url}
\usepackage{etoolbox}
\usepackage{wasysym}
\usepackage[export]{adjustbox}
\usepackage{tikz}
\usepackage{bm}
\usepackage{cite}
\usepackage{academicons}
\usepackage{multirow}
\usepackage{multicol}
\usepackage{algorithm}
\usepackage{algorithmic}

\definecolor{orcidlogocol}{HTML}{A6CE39}
\definecolor{lime}{HTML}{A6CE39}
\DeclareRobustCommand{\orcidicon}{%
    \begin{tikzpicture}
    \draw[lime, fill=lime] (0,0) 
    circle [radius=0.16] 
    node[white] {{\fontfamily{qag}\selectfont \tiny ID}};
    \draw[white, fill=white] (-0.0625,0.095) 
    circle [radius=0.007];
    \end{tikzpicture}
    \hspace{-2mm}
}
\newcommand{\orcidWalter}{\href{https://orcid.org/0000-0003-4565-1272}{\orcidicon}}
\newcommand{\orciddeyu}{\href{https://orcid.org/0009-0006-0905-0540}{\orcidicon}}
\newcommand{\orcidvenkat}{\href{https://orcid.org/0000-0003-1305-2312}{\orcidicon}}
\newcommand{\orcidleah}{\href{https://orcid.org/0000-0001-5433-1452}{\orcidicon}}

\newcommand{\orcidXingcheng}{\href{https://orcid.org/0000-0003-1178-5221}{\orcidicon}}
\newcommand{\orcidMingyu}{\href{https://orcid.org/0000-0002-8752-7950}{\orcidicon}}

\newcommand{\orcidKnoll
}{\href{https://orcid.org/0000-0003-4840-076X}{\orcidicon}}

\makeatletter
\let\NAT@parse\undefined
\makeatother

\usepackage{hyperref}
\hypersetup{
    bookmarks=false,
    colorlinks=true,
    breaklinks=true,
    pagebackref=true,
    linkcolor=blue,
    filecolor=magenta,      
    hypertexnames=true,
    urlcolor=cyan,
    pdftitle={ITSC2024},
    pdfpagemode=FullScreen,
    citecolor=green
    }
\usepackage[capitalize]{cleveref}
\crefname{figure}{Fig.}{Figs.}
\Crefname{figure}{Figure}{Figures}
\crefname{section}{Sec.}{Secs.}
\Crefname{section}{Section}{Sections}
\crefname{table}{Tab.}{Tabs.}
\Crefname{table}{Table}{Tables}
\crefname{equation}{Eq.}{Eqs.}
\Crefname{equation}{Equation}{Equations}

\title{\LARGE \bf
WARM-3D: A Weakly-Supervised Sim2Real Domain Adaptation \\  Framework for Roadside Monocular 3D Object Detection 
}

\author{
Xingcheng Zhou$^{\star\dagger}$\orcidXingcheng \qquad Deyu Fu$^\star$\orciddeyu \qquad Walter Zimmer\orcidWalter \qquad  Mingyu Liu\orcidMingyu  \\ Venkatnarayanan Lakshminarasimhan\orcidvenkat \qquad  Leah Strand\orcidleah \qquad   Alois C. Knoll \orcidKnoll
\thanks{ The authors are with the School of Computation, Information and Technology, Technical University of Munich, 85748 Garching, Germany
}
\thanks{ $^\dagger$ Corresponding Author: \texttt{xingcheng.zhou@tum.de}}
\thanks{ $^\star$ Equal Contribution.}
}

\begin{document}

\makeatletter
\let\@oldmaketitle\@maketitle
\renewcommand{\@maketitle}{\@oldmaketitle
  \centering
  \url{https://WARM-3D.github.io}\\[8pt]
  \setcounter{figure}{0}

  \includegraphics[width=1.0\linewidth]{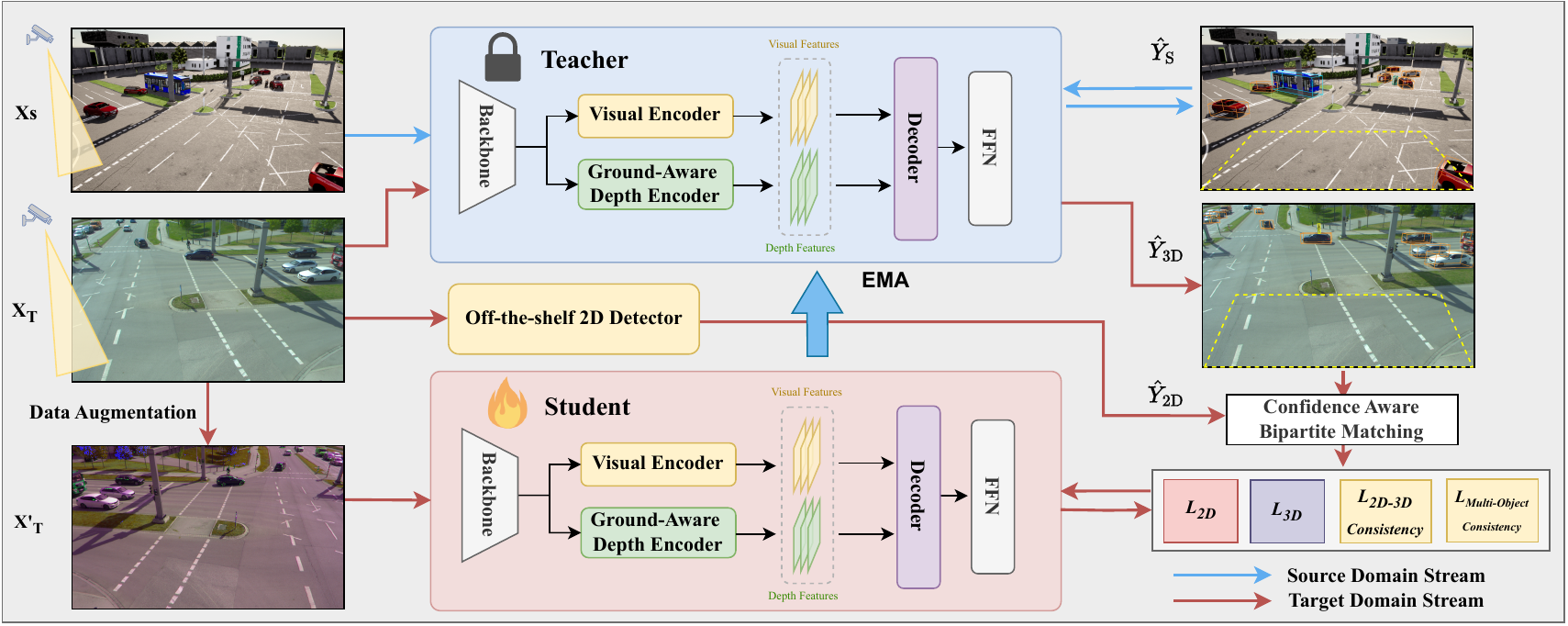}
  \captionof{figure}{Overview of our WARM-3D (\textbf{W}eakly-Supervised Domain \textbf{A}daptation for \textbf{R}oadside \textbf{M}onocular \textbf{3D} Object Detection) Framework. WARM-3D is a cross-domain transfer learning framework for roadside monocular 3D object detection leveraging mature off-the-shelf 2D object detectors to guide the weak-label supervision.}
  \label{fig:title_figure}

  }  
\makeatother

\maketitle





\begin{abstract}

Existing roadside perception systems are limited by the absence of publicly available, large-scale, high-quality 3D datasets. Exploring the use of cost-effective, extensive synthetic datasets offers a viable solution to tackle this challenge and enhance the performance of roadside monocular 3D detection. In this study, we introduce the TUMTraf Synthetic Dataset, offering a diverse and substantial collection of high-quality 3D data to augment scarce real-world datasets. Besides, we present WARM-3D, a concise yet effective framework to aid the Sim2Real domain transfer for roadside monocular 3D detection. Our method leverages cheap synthetic datasets and 2D labels from an off-the-shelf 2D detector for weak supervision. We show that WARM-3D significantly enhances performance, achieving a +12.40\% increase in ${\mathbf{mAP_{3D}}}$ over the baseline with only pseudo-2D supervision. With 2D GT as weak labels, WARM-3D even reaches performance close to the Oracle baseline. Moreover, WARM-3D improves the ability of 3D detectors to unseen sample recognition across various real-world environments, highlighting its potential for practical applications. 

\end{abstract}

\section{Introduction}

With the development of autonomous driving and Vehicle-to-Infrastructure (V2I) technology, 3D perception with intelligent roadside sensors, as an emerging topic, has increasingly gained attention. Compared to vehicle-side, roadside detection exhibits several unique characteristics: First, due to the higher mounting positions of roadside sensors, roadside perception generally has longer detection ranges and is affected by less obstruction. Secondly, roadside sensors are often tilted towards the ground to obtain a better view, which contradicts the assumption of a parallel ground surface in vehicle-side perception and necessitates the consideration of additional pitch and roll angles. Moreover, once installed, the height of road segment sensors rarely changes, which results in a relatively static background environment. This property, to some extent, simplifies the complexity of depth estimation, one of the main challenges faced by vision-based 3D perception. It also causes the performance discrepancy between vision-based and lidar-based approaches in roadside perception to be less apparent than on the vehicle side perception, as attested in \cite{BEVHeight}.

Vision-based 3D object detection relies on annotated 3D bounding boxes as training labels. Labeling precise object positions in 3D space with only one camera, however, presents challenges due to inherent ambiguities. Therefore, prior roadside datasets often employ additional sensors, such as LiDAR, to collect point cloud data for 3D bounding box annotation. The TUMTraf Intersection dataset \cite{TUMTrafIntersection}, for instance, integrates vehicular and roadside LiDAR data for 3D annotation. While being precise, the whole 3D bounding boxes annotation process is costly, time-consuming, and complex, usually involving synchronization, localization, calibration, coordinate transformation, etc. To this end, producing low-cost, extensive synthetic data, especially visual data with high-quality 3D bounding boxes, is a promising method to address these challenges, which can be combined with real-world datasets and serve as a significant supplement. Hence, we introduce the TUMTraf Synthetic dataset (TUMTraf-S), a new roadside dataset designed to promote the development of roadside perception.

We design the WARM-3D framework based on two key observations. Firstly, we notice that the 3D monocular object detector training on the TUMTraf-S retains good pose estimation performance when testing in the real-domain TUMTraf Intersection (TUMTraf-I) dataset \cite{TUMTrafIntersection}, especially in vehicle direction prediction. This suggests that despite the large visual domain gap between simulation and real images, the cross-domain gap for object pose estimation is relatively smaller and can be more easily transferred. Another observation is that even the pre-trained 2D object detectors, such as YOLO series \cite{wang2024yolov9} and DETR series \cite{detr}, outperform fully supervised 3D detectors trained with limited data, especially in out-of-distribution traffic participants recognition. Due to the mature development of 2D object detection and its ease of domain-specific performance enhancement, it is natural to leverage the 2D detectors to assist domain adaption of 3D monocular object detection, to reduce its false positive and false negative samples and improve the quality of domain adaptation. The contributions of this paper are as follows:



\begin{itemize}
  \item We propose WARM-3D, a novel weakly supervised framework to guide domain adaptation of monocular 3D object detection leveraging 2D pseudo labels. 
  \item We present the TUMTraf-S dataset, a large-scale synthetic roadside dataset, and showcase how it benefits real-world roadside 3D object detection.
  \item We introduce Confidence-Aware Bipartite Matching (CABM) to dynamically select high-quality pseudo labels and introduce Multi-Object Coplanar Constraint (MOC) and 2D-3D Projective Consistency (PC) to assist the roadside domain adaptation process. 

  
\end{itemize}

\section{Related Work}



 \begin{figure}[t]
    \centering
      \begin{adjustbox}{width=0.48\textwidth}
        \includegraphics{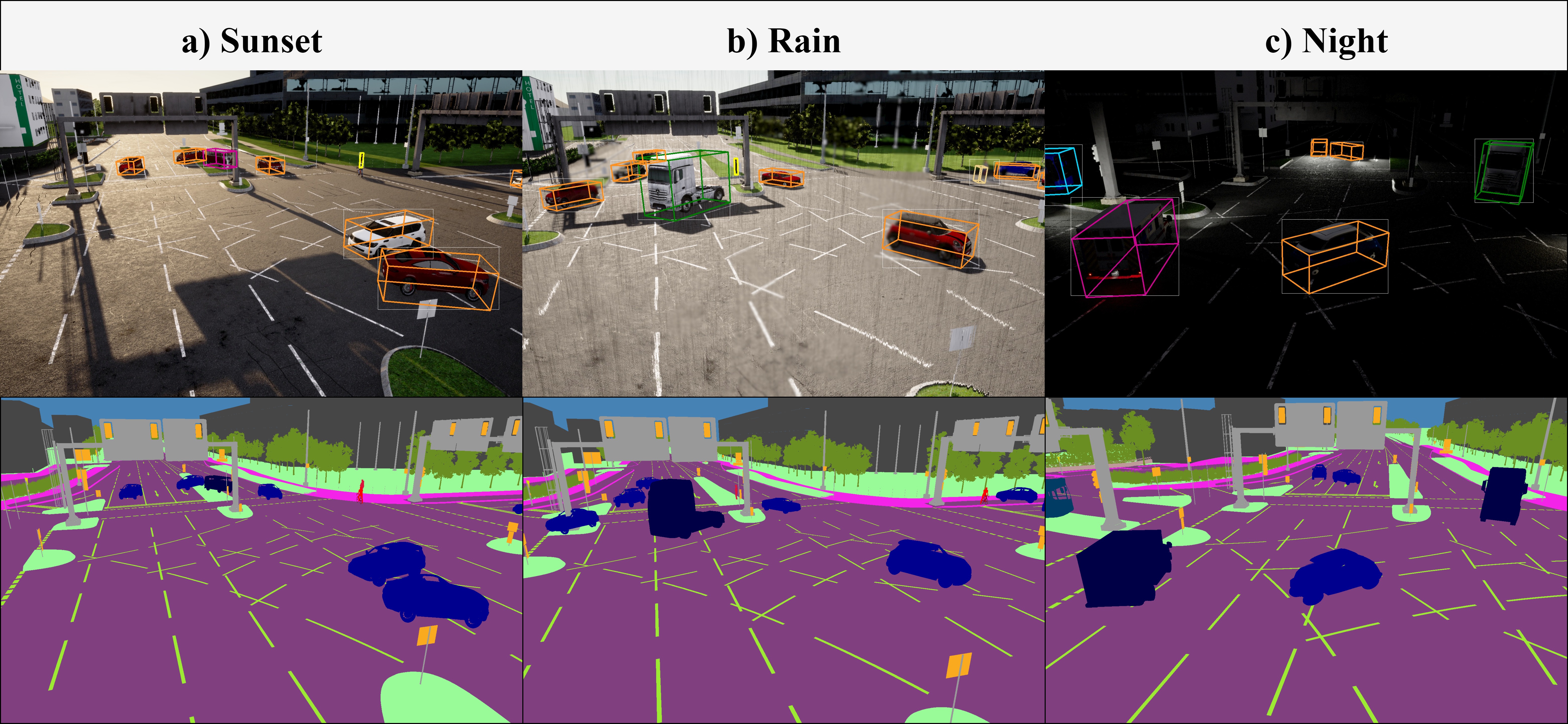}
    \end{adjustbox}

    \caption{Ground truth visualization of TUMTraf-S Dataset. RBG image with 3D bounding box label (top) and instance-level segmentation label (bottom) for three samples: a) sunset, b) rain, and c) night.}
    \label{fig:image_semantic}
    \vspace{-2em}

\end{figure}

\subsection{Monocular 3D Object Detection}
Monocular 3D object detection takes single-view images as input and predicts 3D bounding boxes as output. Due to the lack of depth information, it is inherently ambiguous for monocular 3D object detectors to accurately estimate the positions and sizes of objects within 3D space. Some approaches introduce auxiliary information, such as LiDAR point clouds \cite{pesudolidar3DOD}, to simplify the 3D detection process. While there can be some performance gains with the help of extra input, it usually comes with additional labeling effort and lower inference speed, and also reduces the flexibility of the algorithm.
With the emergence of DETR and its subsequent studies in 2D object detection, some recent methods also apply Transformer to monocular 3D object detection. MonoDTR \cite{monodtr} presents a depth-aware transformer module with the auxiliary supervision of LiDAR during the training process. MonoPGC \cite{MonoPGC} introduces depth-gradient positional encoding to distill rich pixel geometry contexts into the transformer. In this study, we primarily develop our WARM-3D framework based on MonoDETR \cite{monodetr}, a representative transformer-based monocular 3D object detector without additional data. Given the generality and ease of use of the DETR architecture, our framework is be easily adaptable with other DETR-track monocular 3D detectors. 




\subsection{Roadside 3D Object Detection}
As sensors for roadside perception are typically installed at higher elevations, roadside 3D object detectors can overcome the challenges posed by visual occlusion and provide a longer observation distance compared to vehicle-side perception. Some pioneer datasets, such as TUMTraf Dataset series \cite{TUMTrafIntersection, zimmer2024tumtrafv2x}, DAIR-V2X \cite{DAIR-V2X} have been proposed to facilitate the development of roadside perception. Besides, some studies adapt vehicle-based detectors to the characteristics of roadside detection. InfraDet3D \cite{InfraDet3D} designs a multi-modal infrastructure-side perception framework incorporating LiDAR and cameras. \cite{zimmer2023a9} propose a real-time 3d object detector with roadside lidars. MonoUNI \cite{MonoUNI} considers the diversity of pitch angles and focal lengths and proposes a unified network for both infrastructure and vehicle-side detection. Despite the advances, the amount of publicly available roadside 3D data is considerably smaller compared to that for vehicle-based detection \cite{liu2024survey}. Therefore, training a stable and generalized roadside detector capable of handling various environmental conditions in real-world scenarios remains challenging. 


%

\subsection{Weakly-Supervised Domain Adaptation}
For tasks that lack high-quality labeled data but still seek to obtain the practically best model performance, using a more easily available weak label in the target domain for transfer learning is a promising direction. Compared to unsupervised domain adaptation \cite{selftraininng}, weakly supervised domain adaptation allows the use of extra target-domain weak labels. The definitions of weak labels vary across different tasks. For instance, some studies \cite{WeaklySS1,WeaklySS2} take image labels, point labels, bounding boxes, or coarse labels as weakly supervisory labels for semantic segmentation, while \cite{weak-OD} employs image-level annotations from the target domain for weakly-supervised 2D object detection. Some recent studies \cite{jiang2023far3d} employ 2D detectors to assist 3D object detection, which can be viewed as a variant of weak-supervised learning. In our WARM-3D framework, we use instance-level 2D labels obtained from the pre-trained 2D detectors as the weak labels to guide the sim2real cross-domain adaptation process of the monocular 3D object detection.




\begin{figure}[thb]
  \centering
  
  \begin{tabular}{cc}
    \multicolumn{1}{c}{\fontsize{6pt}{5pt}\textbf{TUMTraf Intersection Dataset (TUMTraf-I)}} & \multicolumn{1}{c}{\fontsize{6pt}{5pt}\textbf{TUMTraf Synthetic Dataset (TUMTraf-S)}} \\
  \end{tabular}
  
  \begin{subfigure}[b]{0.48\columnwidth}
    \includegraphics[width=\textwidth]{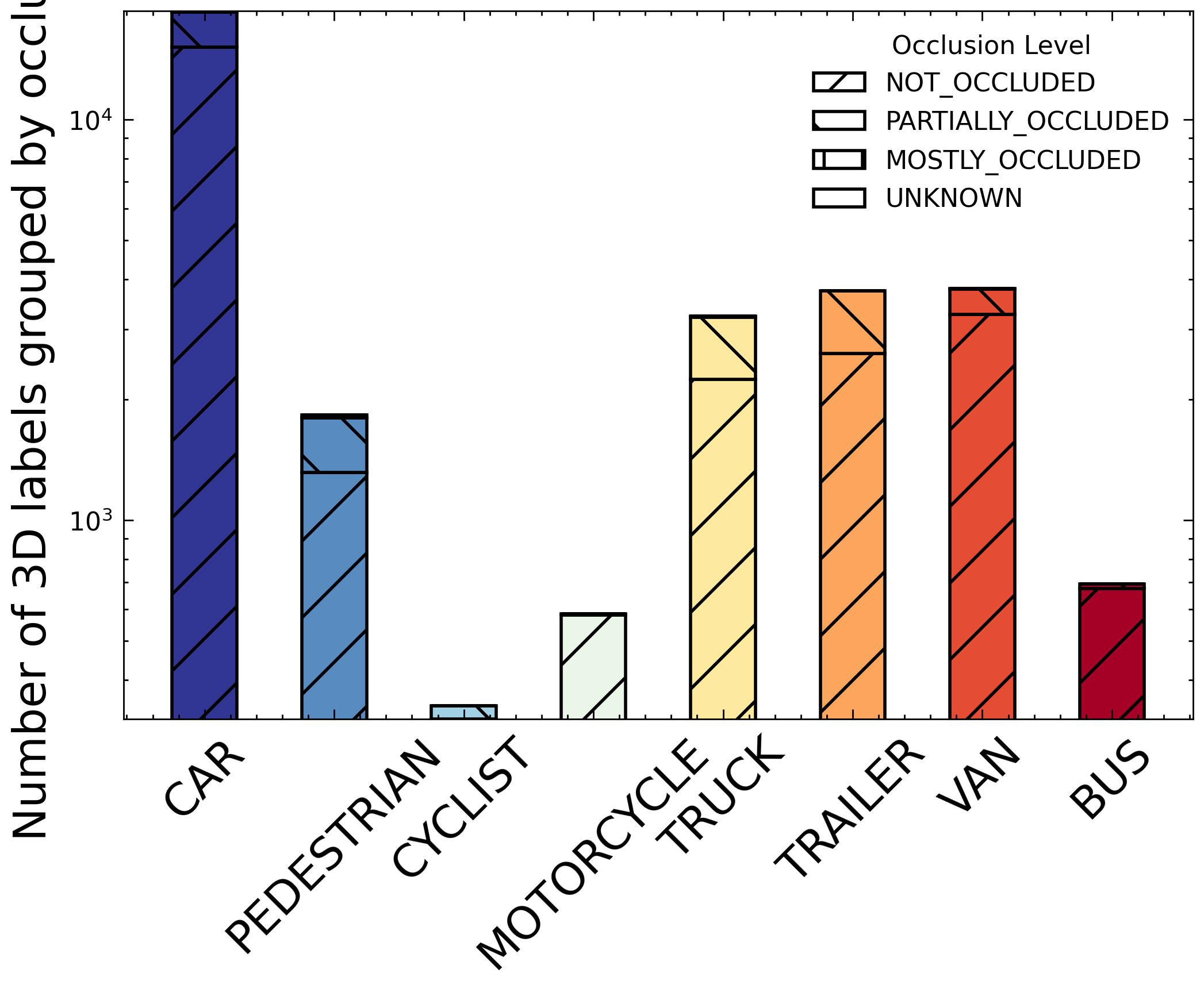}
    \caption{\fontsize{6pt}{5pt}\selectfont Class distribution in TUMTraf-I.} 
  \end{subfigure}
  \hfill 
  \begin{subfigure}[b]{0.48\columnwidth}
    \includegraphics[width=\textwidth]{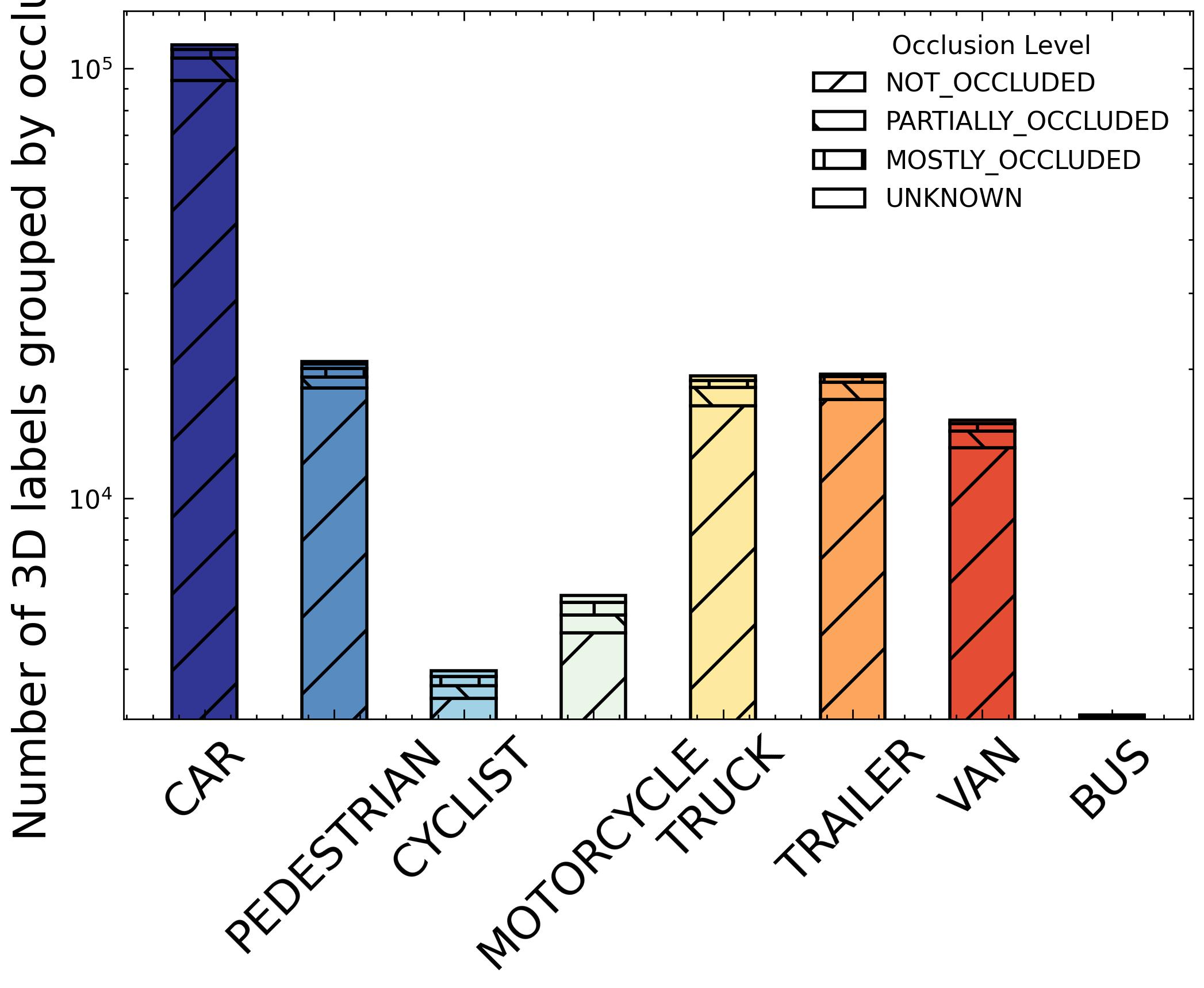}
    \caption{\fontsize{6pt}{5pt}\selectfont Class distribution in TUMTraf-S.} 
  \end{subfigure}

  \begin{subfigure}[b]{0.48\columnwidth}
    \includegraphics[width=\textwidth]{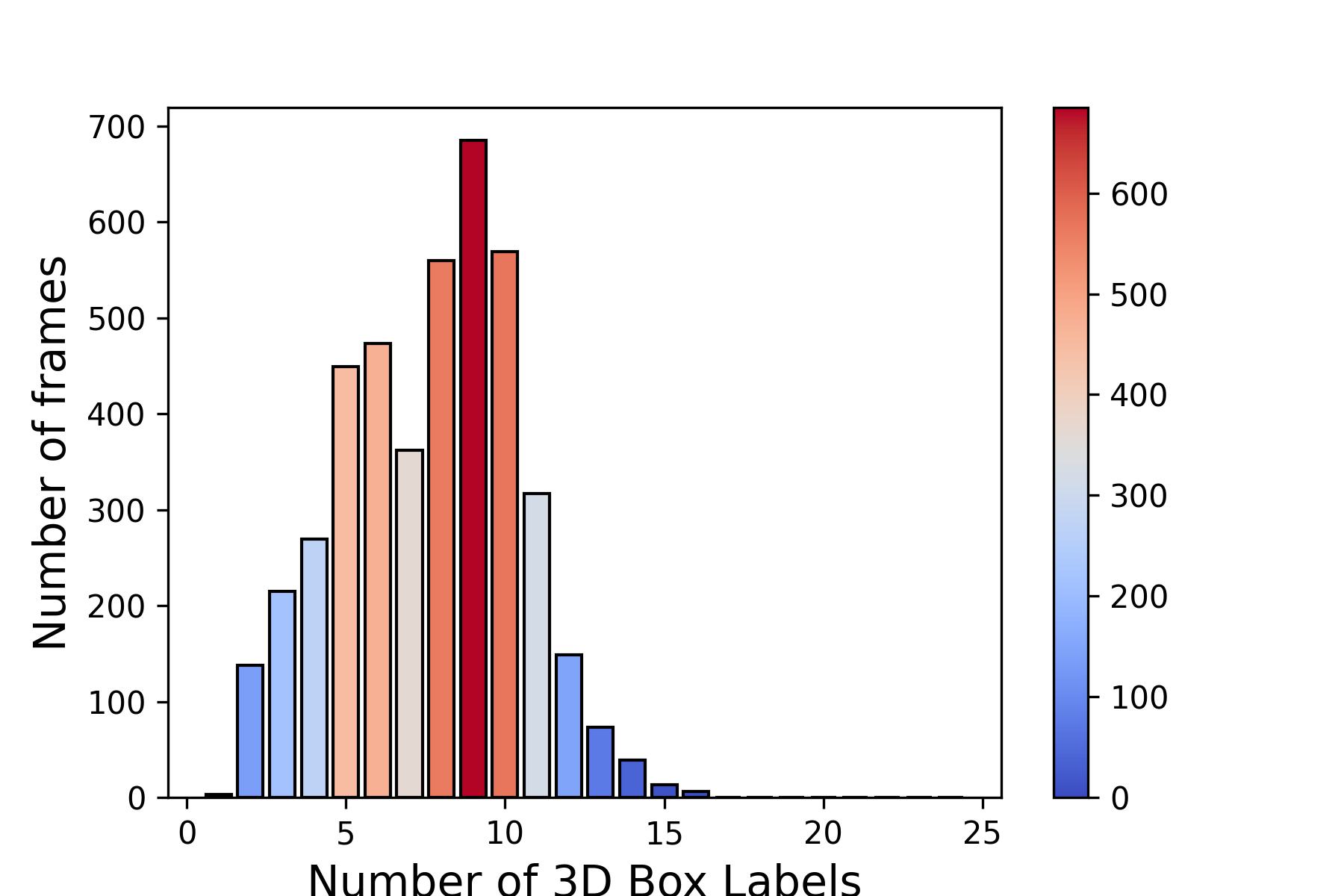}
    \caption{\fontsize{5.5pt}{5pt}\selectfont Num. of 3D labels per frame in TUMTraf-I.} 
  \end{subfigure}
  \hfill
  \begin{subfigure}[b]{0.48\columnwidth}
    \includegraphics[width=\textwidth]{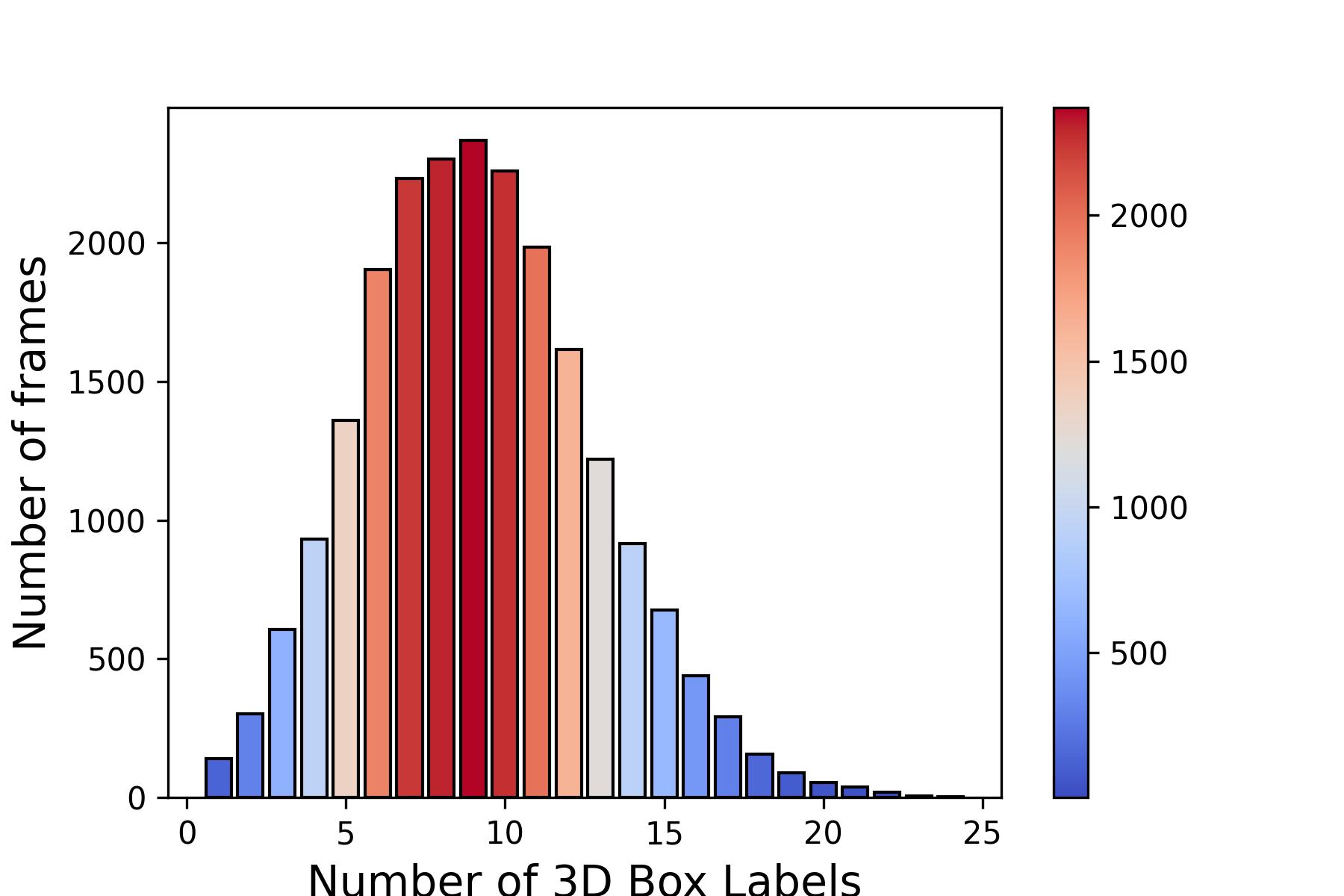}
    \caption{\fontsize{5.5pt}{5pt}\selectfont Num. of 3D labels per frame in TUMTraf-S.} 
  \end{subfigure}

  \begin{subfigure}[b]{0.48\columnwidth}
    \includegraphics[width=\textwidth]{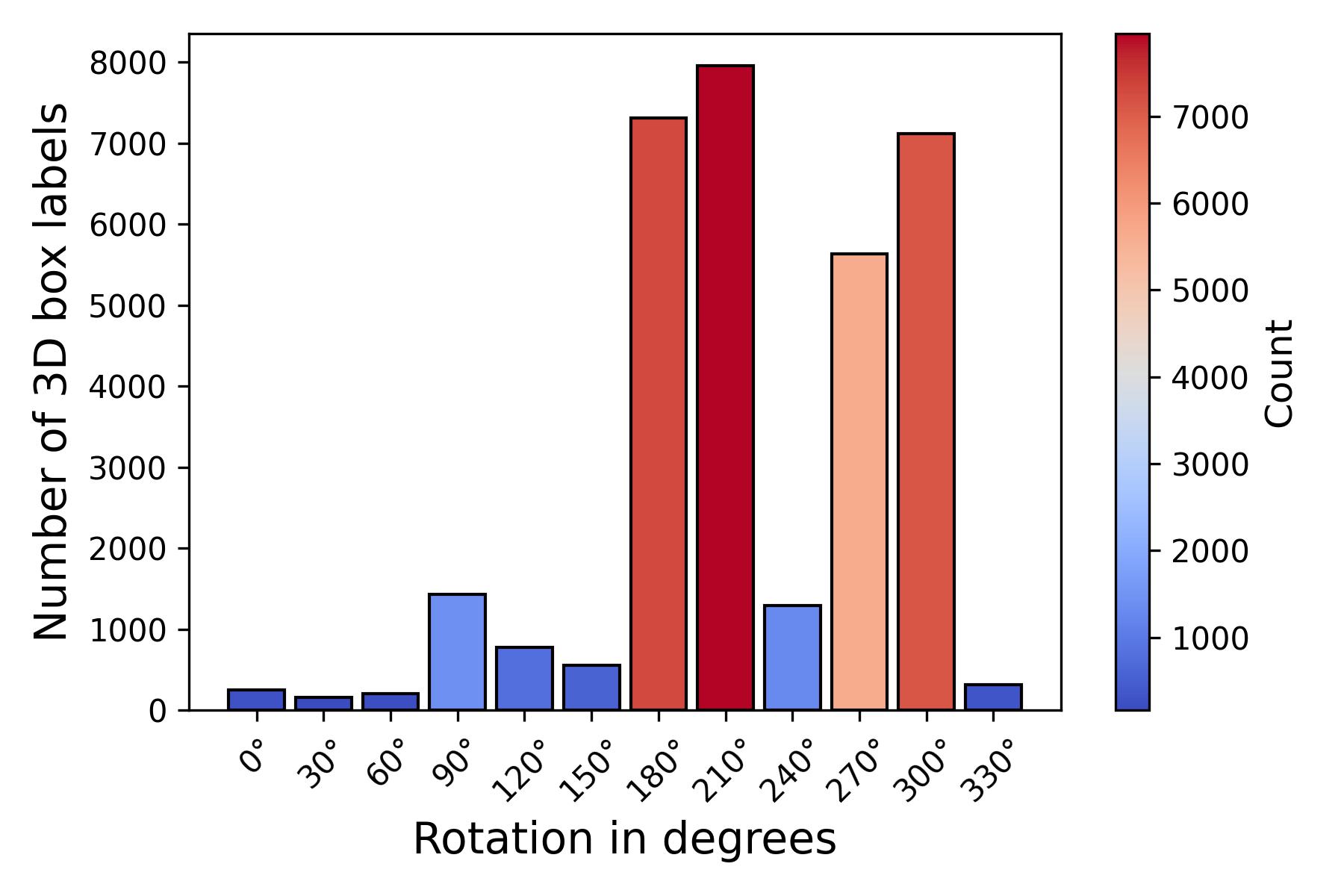}
    \caption{\fontsize{6pt}{5pt}\selectfont Object orientation (yaw) in TUMTraf-I.} 
  \end{subfigure}
  \hfill
  \begin{subfigure}[b]{0.48\columnwidth}
    \includegraphics[width=\textwidth]{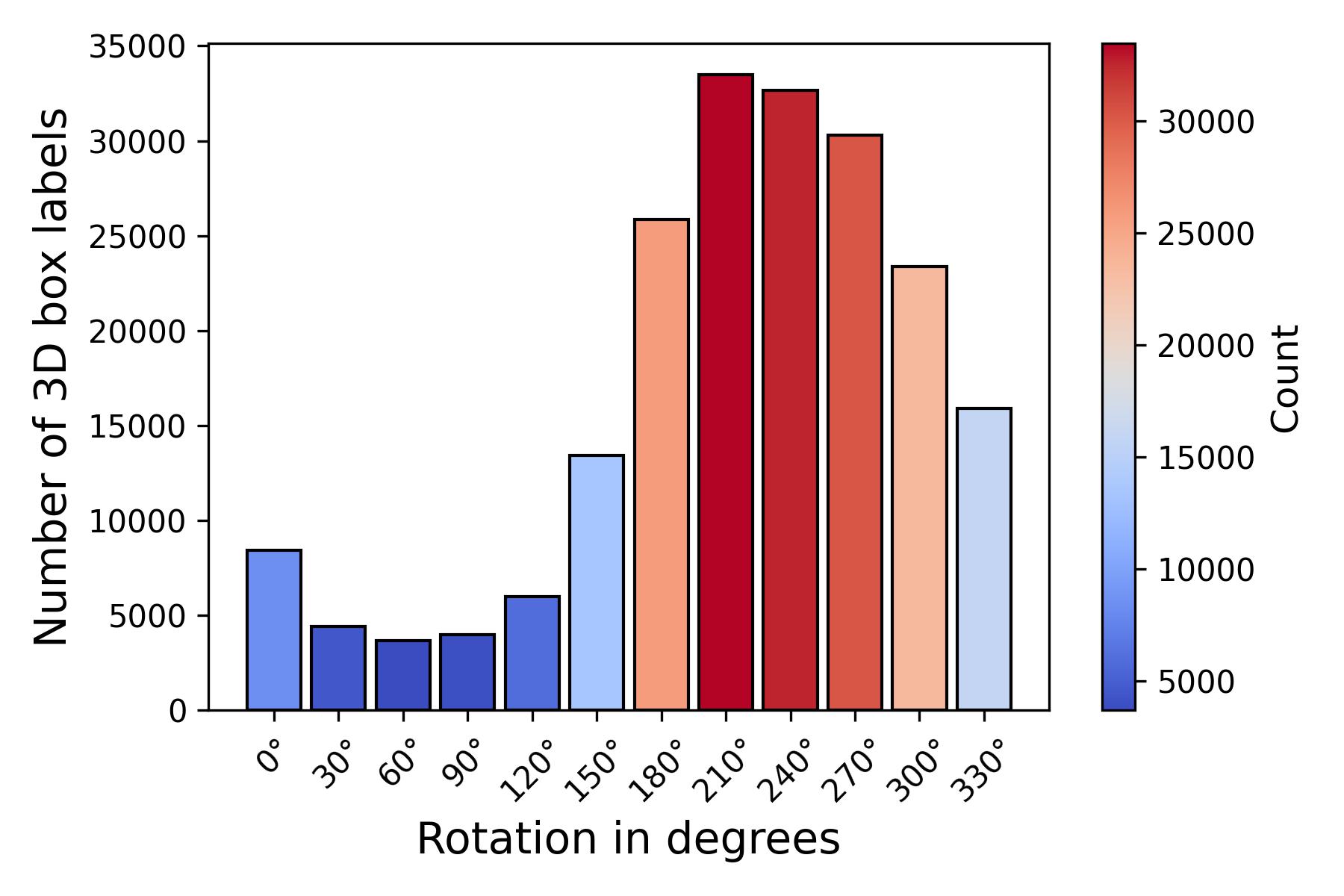}
    \caption{\fontsize{6pt}{5pt}\selectfont Object orientation (yaw) in TUMTraf-S.} 
  \end{subfigure}

  \caption{Comparative analysis of the training set from TUMTraf-I Dataset (left) and the TUMTraf-S Dataset (right), highlighting the data correlations between these two datasets.}
  \label{fig:6figures}
  \vspace{-1.5em}
\end{figure}

\section{TUMTraf Synthetic Dataset}
We introduce the TUMTraf Synthetic Dataset (TUMTraf-S) in this section. The dataset focuses on providing diverse, informative, and large-scale visual data with high-quality 2D and 3D annotations. It greatly expands the real-world TUMTraf-I dataset by delivering 5 times the data quantity.

\subsection{Sensor Setup and Data Capture}
TUMTraf Synthetic Dataset uses the open-sourced CARLA simulator for traffic simulation and data generation. We adopt HD maps as an accurate base source to build the streets and landscapes and create a highly precise 3D world environment map. We adopt RGB cameras as the primary sensor, simulating the data acquisition process found in real-world infrastructure side sensor setups. Unlike real-world TUMTraf-I \cite{TUMTrafIntersection} with fixed view perspective, our TUMTraf-S uniquely manipulates the camera's position and orientation dynamically for each image captured. This random perturbation is designed to generate a more challenging dataset that closely mimics the unpredictability of real-world conditions. Each camera renders images at a resolution of 1920 $\times$ 1080 pixels, providing high-definition visual data essential for detailed analysis and object recognition tasks. 

\subsection{Data Composition and Annotation}
TUMTraf-S dataset contains approximately 24,000 high-resolution images, significantly surpassing the volume of images in real-world TUMTraf Intersection datasets, which is only 4.8k images. These images cover various traffic participants, with great diversity in their appearances, positions, and occlusions. The object poses, sizes, and categories within these images are carefully designed to statistically represent those in real-world datasets, ensuring that simulations are both realistic and varied. The annotations of the TUMTraf-S are formatted in the KITTI dataset format, which includes detailed annotations of various entities. We provide 10 different classes in line with the TUMTraf-I dataset, including Car, Truck, Trailer, Van, Motorcycle, Bus, Pedestrian, Bicycle, etc. The ground truth annotation includes 3D bounding box labels, semantic segmentation maps, and depth maps to provide instance-level and pixel-level labeling information. We show the distribution of the TUMTraf-I and TUMTraf-S datasets in Fig. \ref{fig:6figures}. TUMTraf-S dataset provides an extensive range of simulated weather conditions, especially samples with extreme weather that are hard to capture in real scenarios. We exhaustively configure the environmental factors, including cloudiness, precipitation levels, precipitation deposits, sun azimuth, altitude angles, and fog density. Besides, TUMTraf-S dataset offers a spectrum of lighting conditions ranging from clear, sunny days to dim and murky nights, thus allowing for robustness testing under diverse environmental conditions. 








\section{WARM-3D Framework}



\subsection{Problem Formulation}

We formulate the weakly supervised cross-domain adaptation for roadside monocular 3D object detection problems as follows. Given $N_S$ fully labeled source domain data $\mathcal{D}_S = \{\mathbf{X}
_S^{i},\mathbf{Y}_S^{i},\mathbf{K}_S^{i},\mathbf{E}_S^{i}\}_{i=1}^{N_S}$, where $\mathbf{X}_{S}^{i}$ and $\mathbf{Y}_{S}^{i}$ are source image and label, and $\mathbf{K_{S}}^{i}$, $\mathbf{E_{S}}^{i}$ denote the  camera intrinsic and extrinsic parameters of the $i$-th sample respectively. Besides, there are $N_T$ weakly supervised samples from target domain $\mathcal{D}_T = \{ \mathbf{X}_T^{j}, \mathbf{K}_T^{j},\mathbf{E}_T^{j}\}_{j=1}^{N_T}$ are also available. Moreover, $\mathbf{\hat{Y}}_{T2D}^{j}$ denotes the weak 2D bounding box labels of the $j$-th image from pre-trained 2D detector 
$\mathbf{\hat{Y}}_{2D}^{j} = \mathcal{F}_{2ODet} (\mathbf{X}_T^{j}) $. In this work, we denote the pseudo 2D labels $ \mathbf{\hat{Y}}_{2D}^{j} = \{(c_{2D}^{j}), (x_{2D}^{j}, y_{2D}^{j}), (w_{2D}^{j}, h_{2D}^{j})\}$, in which $c_{2D}^{j}$, ($x_{2D}^{j}$, $y_{2D}^{j}$), ($w_{2D}^{j}$, $h_{2D}^{j}$) mean the class, position and size of the $j$-th 2D bonding box respectively. We denote the 3D bounding box predictions in the target domain as $\mathbf{\hat{Y}}_T$, and its ground truth as $\mathbf{Y}_T$. In roadside 3D object detection, we seek to predict the 3D bounding box with object class $c$, object size $(l, w, h)$, location $(x, y, z)$, and yaw angle $\psi$, with $\mathbf{Y}_T = \{ (c^k),(l^k,w^k,h^k),(x^k,y^k,z^k),(\psi^k) \}_{k=1}^{K} $. $\mathcal{F}_{teacher}$ and $\mathcal{F}_{student}$ are roadside monocular 3D object detector. As a common assumption, we assume the camera's intrinsic and extrinsic parameters in the target domain are available after cam installation. The ultimate goal is to train the model given ${\{ \mathcal{D}_S, \mathcal{D}_T \} }$ that achieve consistently superior 3D predictions $\mathbf{\hat{Y}}_T$ in the target domain.


\newcommand{\ema}[1]{\item[\textbf{EMA Stage:}] #1}
\newcommand{\burnin}[1]{\item[\textbf{Burn-in Stage:}] #1}
\newcommand{\ret}[1]{\item[\textbf{Return:}] #1}

\begin{algorithm}[bth]
\begin{algorithmic}
\REQUIRE ~~\\ $\mathcal{D}_S = \{\mathbf{X}
_S^{i},\mathbf{Y}_S^{i},\mathbf{K}_S^{i},\mathbf{E}_S^{i}\}_{i=1}^{N_S}$,  $\mathcal{D}_T = \{ \mathbf{X}_T^{j},\mathbf{K}_T^{j},\mathbf{E}_T^{j}\}_{j=1}^{N_T}$          
\burnin ~~\\ 
    \small
    $\mathcal{F}_{teacher} = \arg\min \ \textit{L}_{B}  (  \sum \{ \mathbf{Y}_S^{i}, \mathcal{F}_{teacher} (\mathbf{X}_S^{i},\mathbf{K}_S^{i},\mathbf{E}_S^{i}) \}_{i=1}^{N_{S}} ) $

\ema{
\FOR{$ e = 1$ to ${Epoch}$ }
\STATE $\mathbf{\hat{Y}}_{3D} = \mathcal{F}_{teacher} (\mathbf{X}_T) $ \\
\STATE $\mathbf{\hat{Y}}_{2D} = \mathcal{F}_{2ODet} (\mathbf{X}_T) $
\STATE $\{ \mathcal{G}_{3D}, \mathcal{G}_{2D} \} =  \textit{CABM} ( \mathbf{\hat{Y}}_{3D}, \mathbf{\hat{Y}}_{2D} ) $
\STATE $\mathbf{X}^{'}_T = \textit{DataAug} ( \mathbf{X}_T )$
\STATE $ \mathbf{Y}^{'}_{3D} = \mathcal{F}_{student} (\mathbf{X}^{'}_T,\mathbf{K}_T,\mathbf{E}_T) $

\STATE $\mathcal{F}_{student} = \arg\min \ \textit{L}_{EMA}  (  \sum \{ \{ \mathcal{G}_{3D}, \mathcal{G}_{2D} \}, \mathbf{Y}^{'}_{3D} \} ) $
\STATE $\mathcal{F}_{teacher} = \textit{EMA} ( \mathcal{F}_{student}, \mathcal{F}_{teacher}) $ 

\ENDFOR
}

\ret $\mathcal{F}_{student}$, $\mathcal{F}_{teacher}$
\end{algorithmic}
\caption{Working flow of \textbf{WARM-3D} Framework.}
\label{WARM-3D}

\end{algorithm}

\vspace{-1.9em}

\subsection{Framework Overview}

Our WARM-3D framework, as shown in Fig. \ref{fig:title_figure}, is composed of four main components: 1) Confidence-Aware Bipartite Matching, 2) Teacher-Student Mutual Learning, 3) 2D-3D Projective Consistency, and 4) Vehicle Coplanar Constraint. Following most of the domain adaptation frameworks, the training of our framework is first performed with a burn-in stage. We start with training the teacher model in a fully supervised manner on the source domain $ \mathcal{D}_S $ for several epochs, i.e., on the TUMTraf-S Dataset to help generate pseudo-3D labels in the target domain. Then, we freeze the weights of the teacher model and initialize the student model with the same weight. After the burn-in stage, the raw image input $\mathbf{X}_T$ in the target domain is fed to three modules: the frozen teacher model $\mathcal{F}_{teacher}$, the pre-trained 2D detector $\mathcal{F}_{2ODet}$, and through a data augmentation to the student model $\mathcal{F}_{student}$.

The frozen teacher model and the 2D detector then generate pseudo 3D labels $\mathbf{\hat{Y}}_T$, and pseudo 2D labels $\mathbf{\hat{Y}}_{2D}^{j}$ respectively. The Confidence-Aware Bipartite Matching module takes both 3D and 2D pseudo labels and produces a mixed pseudo bounding box set $\{ \mathcal{G}_{3D}, \mathcal{G}_{2D} \}$. This pseudo-bounding box set is then used for the label assignment and weak supervision of the student model update. We introduce two extra losses, namely 2D-3D projective loss and vehicle coplanar loss, to help stabilize the training process of the student model. After the update of the student model parameter through backpropagation and gradient descent, the exponential moving average (EMA) is leveraged to smoothly update the weight of the teacher model. We formulate the WARM-3D training process in Alg. \ref{WARM-3D}.



\begin{figure}[thb]
    \centering
   \includegraphics[width=0.47\textwidth]{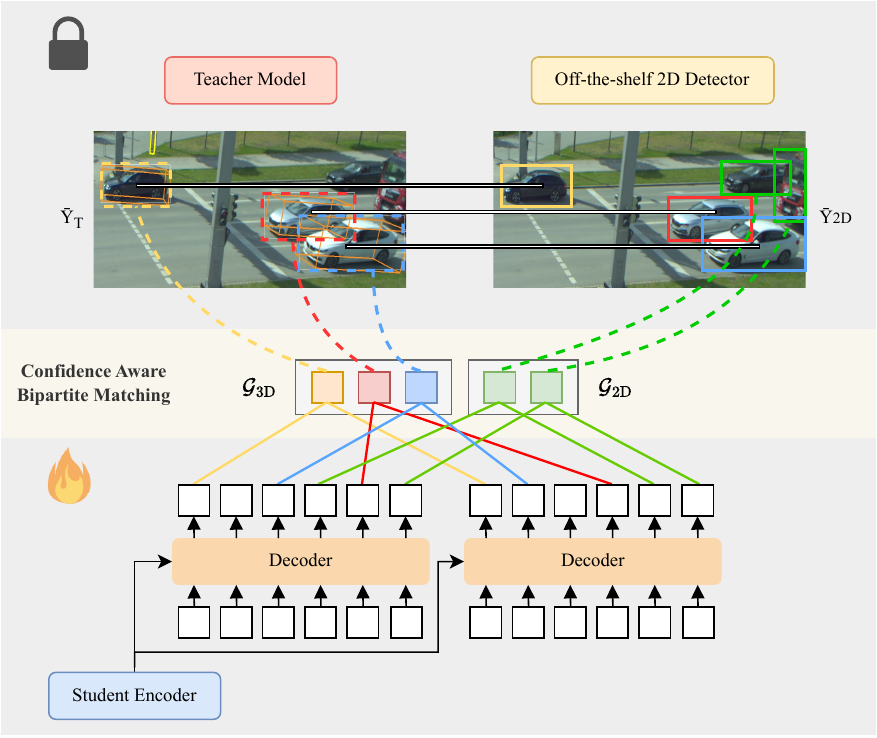}
    \caption{Illustration of the confidence-aware bipartite matching process. $\{ \mathcal{G}_{3D}, \mathcal{G}_{2D} \}$ denotes the matched 3D bounding box set and kept 2D bounding box set. The lock and fire represent the frozen and unfrozen models, respectively.}
    \label{fig:dynamic_macthing}
    \vspace{-1em}
\end{figure}

\begin{figure*}[thb]
    \centering
   \includegraphics[width=0.99\textwidth]{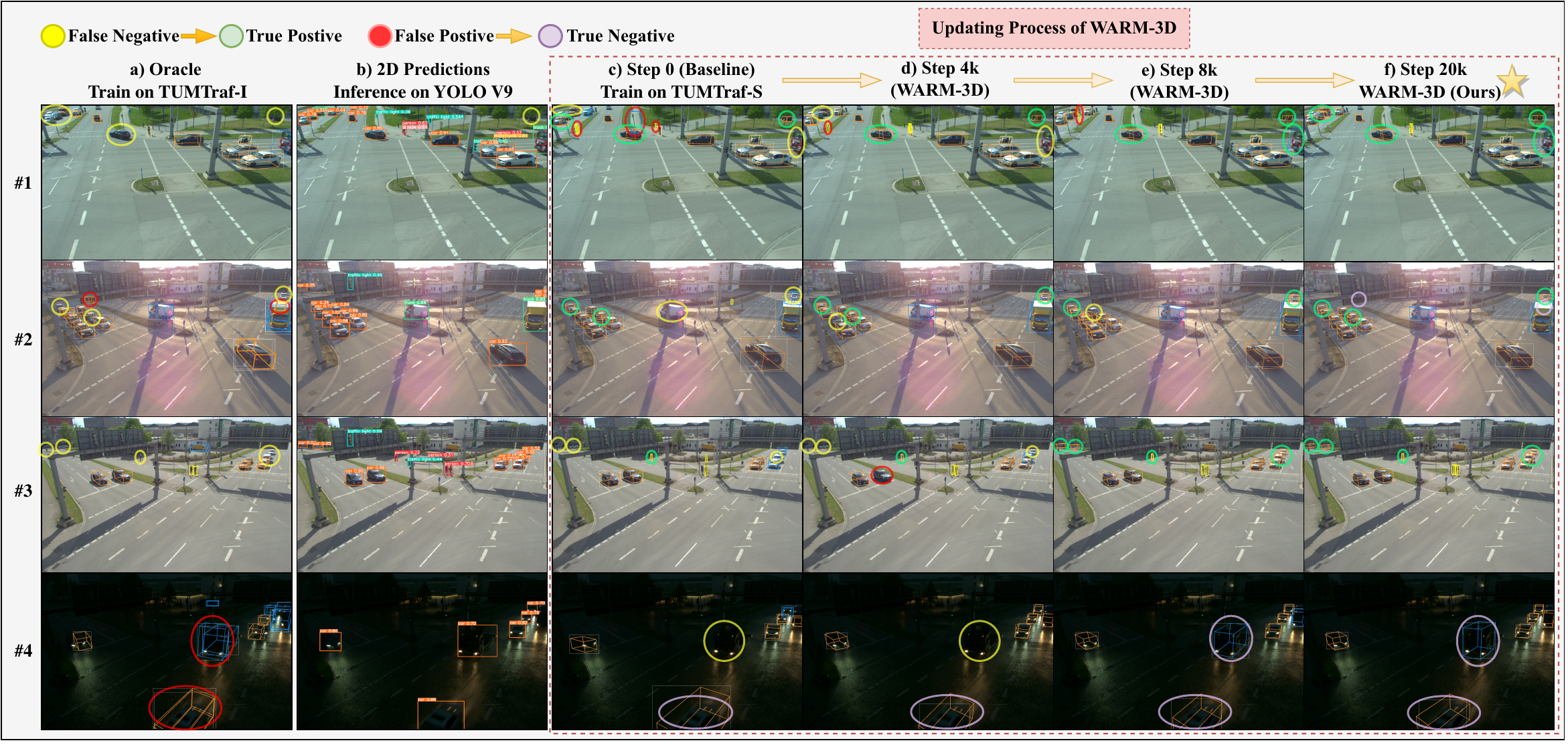}
    \caption{Visualizations of WARM-3D performance on TUMTraf-I Dataset test set. The WARM-3D in f) method performs even qualitatively better than the oracle method in a). The WARM-3D adaptation process is shown from 0 training steps in c) to 20k training steps in f). It's evident that the \textcolor{yellow}{False Negative} objects turn to \textcolor{green}{True Positive}, while \textcolor{red}{False Positive} objects become \textcolor{violet}{True Negative} in the process. }
    \label{fig:vis_result_a9}
\vspace{-1.5em}
\end{figure*}

\subsection{Confidence-Aware Bipartite Matching}
In the domain adaptation process, the teacher model usually produces higher-quality pseudo-3D labels with weakly augmented input to guide the student model. However, it's still hard for both the teacher and student model to be aware of the real quality of these pseudo labels. To this end, we propose to introduce an off-the-shelf 2D detector as an extra indicator to guide the label assignment process and assist the cross-domain adaptation. As illustrated in Fig. \ref{fig:dynamic_macthing}, given 3D predictions from teacher model $\mathbf{\hat{Y}}_{3D} = \{ \hat{y}_{3D}^{i} \}_{i=1}^{N}  $ and 2D predictions from pre-trained 2D detector $\mathbf{\hat{Y}}_{2D} = \{ \hat{y}_{2D}^{j} \}_{j=1}^{M} $, we first project the 3D bounding boxes to 2D images to obtain the corresponding projective 2D bounding boxes $\mathbf{\hat{Y}}^{3D}_{2D}$. Next, we compute the bipartite matching cost between projected 2D bounding boxes $\mathbf{\hat{Y}}^{3D}_{2D}$ and the pseudo 2D bounding box $\mathbf{\hat{Y}}_{2D}$ utilizing the Hungarian algorithm. To take the self-aware box quality into consideration, we also include the prediction confidence to calculate the matching cost. We formulate the cost and matching function as follows in Eq. \ref{eq1} and Eq. \ref{eq2}:

\begin{equation}
    \textbf{C}_{\text{matching}} = \lambda_{\text{class}} \cdot \mathcal{L}_{\text{class}} + \lambda_{\text{giou}} \cdot \mathcal{L}_{\text{giou}} + \lambda_{\text{conf}} \cdot \mathcal{L}_{\text{conf}}
\label{eq1}
\end{equation}

\begin{equation}
    \small
    \{ \mathcal{G}_{3D}, \mathcal{G}_{2D} \} = \underset{\mathbf{g}_{3D}, \mathbf{g}_{2D} \in \{ \mathbf{\hat{Y}}_{3D}, \mathbf{\hat{Y}}_{2D} \}}{\arg\min} \sum_{i} \sum_{j} \textbf{C}_{\text{matching}}(\hat{y}^{3Di}_{2D}, \hat{y}_{2D}^{j})
\label{eq2}
\end{equation}

\subsection{Geometric Constraints}
Due to the underlying inconsistency between the input 2D image and output 3D prediction, monocular 3D detectors are usually difficult to obtain accurate 3D results. Therefore, it is common to utilize geometric constraints to improve the performance. In this study, we introduce two reasonable geometric assumptions well-suited for vision-based roadside detection to fully utilize the pseudo labels of confidence-aware bipartite matching. \\

\textbf{2D-3D Projective Consistency.}
The presence of the camera's pitch and roll angles in roadside perception usually brings the benefit of less view occlusion. We observe that in this case, the projected 2D bounding box abstracted from 3D predictions usually closely overlaps with the 2D detection pseudo box $\mathbf{\hat{Y}}_{2D}$. Therefore, we assume the projected 2D bounding box from 3D predictions should be softly close to $\mathbf{\hat{Y}}_{2D}$. Given 3D predictions from the student model, $ \mathbf{Y}^{'}_{3D} = \mathcal{F}_{student} (\mathbf{X}^{'}_T,\mathbf{K}_T,\mathbf{E}_T) $, we first project the 8 corners of 3D bounding boxes from the object coordinate to the image coordinate, as defined in Eq. \ref{eq:3} and Eq. \ref{eq:4}, where $\textit{Cor}_{obj}^{3D}$ denotes the homogeneous format of 8 corners in object coordinate 
 $ \textit{Cor}_{obj}^{3D} \in \mathbb{R}^{4 \times 8} $, and $T$ denotes the transformation vector. The $yaw$ angle $\psi$ is from $\mathbf{Y}^{'}_{3D}$, $pitch$ $\theta$ and $roll$ $\phi$ are extracted from the ground plane. 

\begin{equation}
\begin{gathered}
    \textit{Cor}_{img}^{2D} = K \left[ \begin{array}{cc}
    R_x @ R_y @ R_z & T \\
    \mathbf{0}^T & 1
    \end{array} \right] \textit{Cor}_{obj}^{3D} \\
\end{gathered}
\label{eq:3}
\end{equation}
\begin{equation}
\small 
\setlength{\arraycolsep}{0.3pt}
\begin{gathered}
    R_x @ R_y @ R_z = \\ \begin{bmatrix}
    1 & 0 & 0 \\
    0 & \cos(\phi) & -\sin(\phi) \\
    0 & \sin(\phi) & \cos(\phi)
    \end{bmatrix}  \begin{bmatrix}
    \cos(\theta) & 0 & \sin(\theta) \\
    0 & 1 & 0 \\
    -\sin(\theta) & 0 & \cos(\theta)
    \end{bmatrix}  \begin{bmatrix}
    \cos(\psi) & -\sin(\psi) & 0 \\
    \sin(\psi) & \cos(\psi) & 0 \\
    0 & 0 & 1
    \end{bmatrix} 
\end{gathered}
\label{eq:4}
\end{equation}

We calculate the projected 2D bounding box $\mathbf{Y}^{'}_{2D} $ from $\textit{Cor}_{img}^{2D}$, and adopt the projective consistency loss in Eq. \ref{eq:projectiveconsistency}. We apply GIoU loss for $\mathcal{L}_{giou}$ and L1 loss for $\mathcal{L}_{center}$.

\begin{equation}
    \mathcal{L}_{\text{2D-3D Consistency}} = \lambda_{\text{giou}} \cdot \mathcal{L}_{\text{giou}} + \lambda_{\text{center}} \cdot \mathcal{L}_{\text{center}} 
\label{eq:projectiveconsistency}
\end{equation}


\begin{table*}[thb!]
\centering
\footnotesize 
\caption{Performance of baseline model and WARM-3D framework on TUMTraf-I Dataset test set. We report the $mAP_{3D@0.1}$ results following the settings in \cite{InfraDet3D}. The WARM-3D with 2D ground truth as weak labels shows almost close performance to the Oracle baseline.} 
\resizebox{\textwidth}{!}{
    \begin{tabular}{cccccccccccc}
      \toprule[0.9pt]
      \multirow{2}{*}{\textbf{Method}} & \multicolumn{4}{c}{\textbf{Training Set}} & \multirow{2}{*}{\textbf{Precision}} & \multirow{2}{*}{\textbf{Recall}} & \multicolumn{4}{c}{$\textbf{mAP}_{\textbf{3D}}$}                                                                                                                                                                         \\
      ~                                & TUMTraf-I LiDAR                           & TUMTraf-I Image                     & TUMTraf-S                        & ~                                                & ~                         & ~                         & Easy                      & Mod.                      & Hard                      & Overall                   \\

      \midrule[0.5pt]

      PointPillars \cite{InfraDet3D}   & \checkmark                                & -                                   & -                                &                                                  & 55.23                     & 77.53                     & -                         & -                         & -                         & 54.68                     \\
      MonoDet3D \cite{InfraDet3D}      & -                                         & \checkmark                          & -                                &                                                  & 30.05                     & 36.65                     & -                         & -                         & -                         & 29.69                     \\
      InfraDet3D \cite{InfraDet3D}     & \checkmark                                & \checkmark                          & -                                &                                                  & 55.53                     & 65.39                     & -                         & -                         & -                         & \textbf{55.07}            \\
      Baseline (Oracle)                & -                                         & \checkmark                          & -                                &                                                  & 50.91                     & 55.76                     & 30.31                     & 54.74                     & 49.96                     & {50.57}                   \\
      \midrule[0.5pt]
      Baseline (Source only)           & -                                         & -                                   & \checkmark                       &                                                  & 29.65                     & 29.44                     & 40.52                     & 29.93                     & 14.66                     & 29.15                     \\
      \midrule[0.5pt]

      WARM-3D (YOLOv9)                 & -                                         & -                                   & \checkmark                       &                                                  & 42.23                     & 39.84                     & 55.09                     & 35.17                     & 24.89                     & \textbf{41.55}            \\ \rowcolor{gray!20}

      Improvement                      &                                           &                                     &                                  &                                                  & \textcolor{blue}{+12.57} & \textcolor{blue}{+10.36} & \textcolor{blue}{+14.57} & \textcolor{blue}{+5.25}  & \textcolor{blue}{+10.24} & \textcolor{blue}{+12.40} \\
      \midrule[0.5pt]

      WARM-3D (2D GT)                  & -                                         & \checkmark                          & -                                &                                                  & 48.49                     & 48.04                     & 50.19                     & 51.078                    & 33.565                    & \textbf{48.09}            \\
      \rowcolor{gray!20}
      Improvement                      &                                           &                                     &                                  &                                                  & \textcolor{blue}{+18.84} & \textcolor{blue}{+18.60} & \textcolor{blue}{+9.67}  & \textcolor{blue}{+21.15} & \textcolor{blue}{+18.91} & \textcolor{blue}{+18.94} \\

      \bottomrule[0.5pt]
    \end{tabular}
}

\label{tab:DA-RM3D}
\vspace{-1.2em}

\end{table*}

\textbf{Multi-Objects Coplanar Constraints.}
Besides, we also adopt another reasonable assumption to further stabilize the domain adaptation process. In vehicle-centric monocular 3D object detection, it is common to utilize the ground plane to enhance the depth estimation ability. However, in roadside perception, despite the fixed camera mounting height, roadside units are still susceptible to swaying, leading to variants and bias in the camera's roll and pitch angles and then harming the perception performance. To this end, we adopt a relaxed version plane assumption rather than strictly adhering to the ego-centric ground plane constraint. We only assume that all objects' bottom centers are coplanar rather than necessarily on a given plane that might be subject to disturbance errors. This constraint is unaffected by camera disturbances, making it more adaptable for practical cases. 
\begin{equation}
\mathcal{L}_{\text{MOC}} = \frac{1}{M} \sum_{i=1}^{M} \mathbb{I}(N_i \geq 4)  \cdot  V_{i}[2]
\label{eq:moc}
\end{equation}
In implementation, we use principal component analysis (PCA) to fit the position of the bottom center of all objects $ \textit{X} \in \mathbb{R}^{N \times 3} $ in the plane, where $N$ is the number of objects. We compute the principal component through the SVD decomposition and normalize the variance across three dimensions. We take the last channel of normalized variance $V[2]$ as the coplanar loss $\mathcal{L}_{\text{MOC}}$. As shown in \ref{eq:moc}, it will only be considered when $N \geq 4$, in which \text{MOC} means multiple objects consistency, and \text{M} denotes the number of images.

\subsection{Overall Loss}
The overall loss contains five parts. Following \cite{monodetr}, we disentangle the detection loss into 2D loss $ \mathcal{L}_{\text{2D}} $ and 3D loss  $ \mathcal{L}_{\text{3D}} $ separately. $ \mathbb{M}_{3D} $ denotes the mask to differentiate the objects with 3D information in $\{ \mathcal{G}_{3D}, \mathcal{G}_{2D} \}$ from Confidence-Aware Bipartite Matching. We adopt $\mathcal{L}_{\text{dmap}}$ in \cite{monodetr} without its geometric depth loss. $\mathcal{L}_{\text{2D-3D PC}}$ and $\mathcal{L}_{\text{MOC}}$ denote 2D-3D projective consistency loss and multi-objects coplanar loss respectively. We formulate the overall loss of WARM-3D framework in the EMA stage as follows:

\begin{equation}
\begin{split}
     \mathcal{L}_{\text{Overall}} = &  \lambda_{\text{2D}} \cdot \mathcal{L}_{\text{2D}} + \lambda_{\text{3D}} \cdot \mathbb{M}_{3D} \cdot \mathcal{L}_{\text{3D}} +  \lambda_{\text{dmap}} \cdot \mathcal{L}_{\text{dmap}} \\
    & + \lambda_{\text{2D-3D PC}} \cdot \mathcal{L}_{\text{2D-3D PC}} + \lambda_{\text{MOC}} \cdot \mathcal{L}_{\text{MOC}}
\end{split}
\label{eq:overall}
\end{equation}




\section{Experiments}

\subsection{Settings}
\textbf{Datasets and Evaluation Metrics.}
We evaluate the WARM-3D framework using the TUMTraf Intersection Dataset \cite{TUMTrafIntersection}, which comprises a total of 4,800 images captured from A9 intersections. We follow the dataset settings in \cite{zimmer2023a9} to split the dataset into training, validation, and testing sets with ratios of 80\%, 10\%, and 10\%, respectively. The performance of the WARM-3D framework is assessed using standard 3D object detection metrics: Precision, Recall, and 3D mean Average Precision (${mAP}_{{3D}}$). To provide a comprehensive evaluation, we follow \cite{TUMTrafIntersection} and report detection results across three difficulty levels - easy, moderate, and hard.  


\textbf{Implementation Details.}
We adopt MonoDETR\cite{monodetr}, a center-guided DETR-based monocular 3D detector, with simple modifications as our baseline model. We adopt a ResNet-50 as the backbone following the settings of MonoDETR while removing the geometric depth in the depth prediction head to adapt the roadside scenario. We use the AdamW optimizer, with an initial learning rate of $1 \times 10^{-4}$, and weight decay $1 \times 10^{-4}$. The detection range of the model extends up to 120 meters, with the input images resized to a quarter of their original dimensions, resulting in a resolution of $960 \times 600$ pixels. We map the dataset class into Car, Pedestrian, Cyclist, and Big Vehicles during the training process. The hyperparameter $\lambda_{\text{2D}}$ includes coefficients of labels classification loss and bounding box loss, while $\lambda_{\text{2D}}$ incorporates depth loss, 3d center loss, dimension loss, orientation loss, and depth map loss.

We train the teacher model for 50 epochs in the teacher model burn-in stage and train with WARM-3D framework for 20,000 extra steps. The momentum parameter in the EMA process is set as 0.999, and the update interval is set to 200 steps. We adopt some commonly used data argumentation strategies, including flipping, cropping, image distortion, affine transformation, etc., for teacher-student mutual learning in the WARM-3D framework. We transform inputs of the student model with strong augmentation while choosing weak augmentation for the teacher model. All experiments are conducted on one NVIDIA RTX 3090, with batch size 8.

\begin{figure*}[t]
    \centering
    \includegraphics[width=0.99\textwidth]{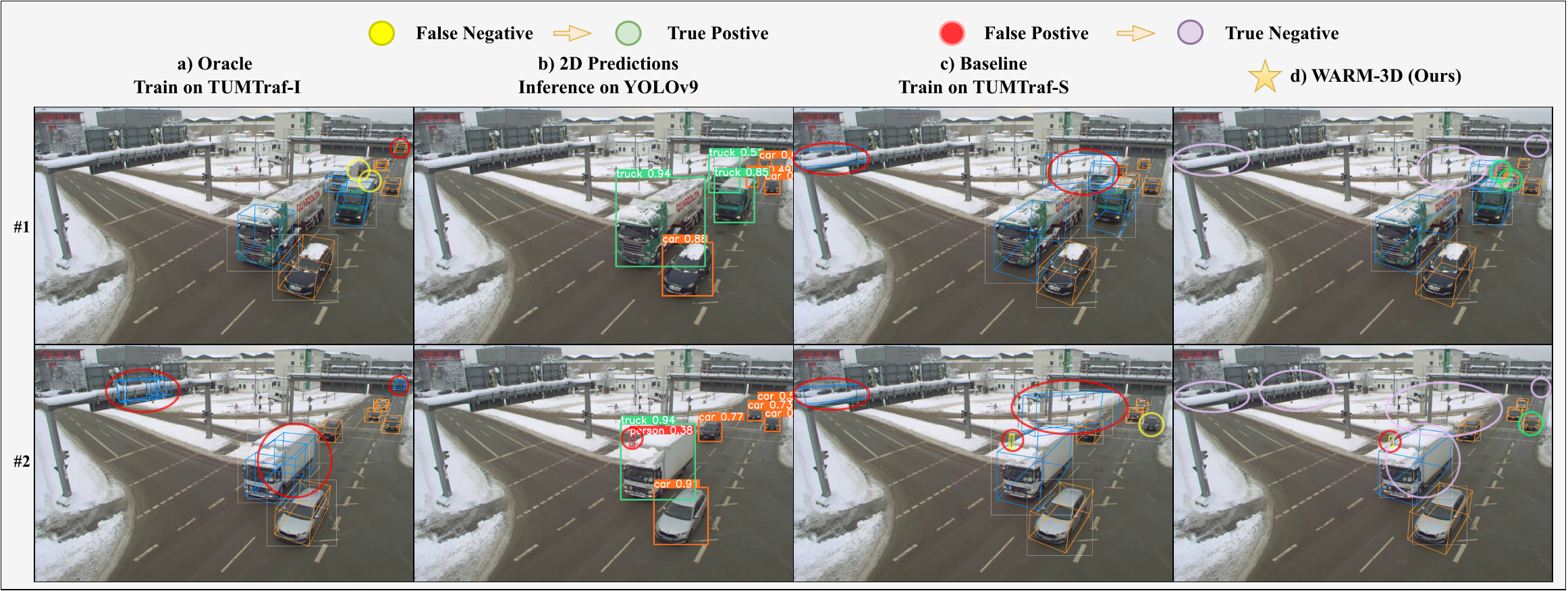}
    \caption{Qualitative results of WARM-3D on unseen samples (outside TUMTraf-I Dataset) in snow conditions. The first column a) shows the results of the Oracle model, and the last column d) visualizes the inference results with our WARM-3D framework, which demonstrates robustness to out-of-distribution scenarios.}
    \label{fig:vis_da_rm3d}
    \vspace{-1em}

\end{figure*}

\subsection{Comparison Methods}
We compare the proposed WARM-3D framework against prior studies and baseline methods to assess its performance in various settings. All the models are evaluated on the test set of the TUMTraf Intersection dataset, and the experiment settings are as follows:

\textbf{Baseline (Oracle):} The model is trained with 3D ground truth labels in the Image set of the TUMTraf-I dataset in a fully supervised manner.

\textbf{RM3D (Source Only):} The model is trained with 3D labels on the TUMTraf-S Synthetic dataset without any domain adaptation process.

\textbf{WARM-3D (YOLOv9):} Model adopts RM3D (Source only) as teacher model and employs WARM-3D framework with 2D pesudo labels from YOLOv9. 

\textbf{WARM-3D (2D GT):} Model adopts RM3D (Source only) as teacher model and employs WARM-3D framework with 2D labels from TUMTraf-I dataset.


\subsection{Effectiveness of WARM-3D Framework}
The effectiveness of the WARM-3D framework is evident compared with the baseline (Source only) model. The baseline (Source Only) model achieved a $\textit{mAP}_{3D}$ of 29.15\% across all difficulty levels. This model particularly struggles in the \textit{hard} category, where the $\textit{mAP}_{3D}$ was only 14.66\%, indicating significant difficulty detecting objects under complex or heavily occluded conditions. In contrast, the WARM-3D (YOLOv9) model, which incorporates domain adaptation techniques, significantly improved performance, with an overall ${mAP}_{{3D}}$ of 41.55\%. This represents a 12.40\% increase in ${mAP}_{{3D}}$ overall compared to the RM3D (Source Only) model. These results underscore the capability of the WARM-3D framework to effectively bridge the gap between the synthetic domain and the real-world domain. The results are summarized in the Table \ref{tab:DA-RM3D}.

\textbf{Qualitative Results.}
Fig. \ref{fig:vis_result_a9} presents the detection results on the TUMTraf-I test set, emphasizing the enhancements in reducing both false positives and false negatives, significantly improving the model's accuracy and reliability. Furthermore, Fig. \ref{fig:vis_da_rm3d} displays the WARM-3D framework's performance on unseen samples under snow conditions, illustrating its adaptability and resilience in handling out-of-distribution scenarios. These visualizations validate the quantitative improvements and showcase the practical effectiveness of the framework in real-world conditions.

\subsection{Ablation Studies}

\begin{table}[htb!]
\centering
\footnotesize 
\caption{Impact of specific elements of the WARM-3D framework on results tested on TUM-Traf-I Image Dataset. CABM refers to Confidence-Aware Bipartite Matching. PC represents 2D-3D Projective Consistency. MOC stands for Multi-Objects Coplanar Constraints.} 

\resizebox{0.48\textwidth}{!}{







\begin{tabular}{cccccccccc}
      \toprule[0.5pt]
      \multicolumn{3}{c}{\textbf{Sim $\rightarrow$ Real}} & \multirow{2}{*}{\textbf{Precision}} & \multirow{2}{*}{\textbf{Recall}} & \multirow{2}{*}{$\textbf{mAP}_{\textbf{3D}} (Overall) $ } & \multirow{2}{*}{\textbf{Improvements}}                                                        \\
      CABM                                                & PC                                  & MOC                              & ~                                                         & ~                                      &                &                                   & \\
      \midrule[0.5pt]

      -                                                   & -                                   & -                                & 38.16                                                     & 36.73                                  & 35.32          & -                                   \\

      \checkmark                                          & -                                   & -                                & 38.16                                                     & 38.87                                  & 37.45          & \textcolor{blue}{+2.13}            \\

      -                                                   & \checkmark                          & -                                & 39.07                                                     & 39.02                                  & 38.36          & \textcolor{blue}{+3.04}            \\

      -                                                   & -                                   & \checkmark                       & 38.26                                                     & 39.71                                  & 37.58          & \textcolor{blue}{+2.26}            \\

      \checkmark                                          & \checkmark                          & \checkmark                       & \textbf{42.23}                                                     & \textbf{39.84}                                  & \textbf{41.55} & \textbf{\textcolor{blue}{+6.23}}   \\

      \bottomrule[0.5pt]
    \end{tabular}
}
\label{tab:Ablation-DA-RM3D}
\vspace{-1em}

\end{table}









\textbf{Component Impact of WARM-3D framework.}
We explore the impact of individual components of the WARM-3D framework on the overall performance by comparing a baseline model with domain adaptation but without any component mentioned above, shown in Table \ref{tab:Ablation-DA-RM3D}. Introducing Confidence-Aware Bipartite Matching (CABM) alone leads to an increase in overall ${mAP}_{{3D}}$ to 37.45\%, demonstrating the effectiveness of confidence-aware strategies in managing object detection uncertainties. Projective Consistency (PC) further enhanced model performance, bringing the overall ${mAP}_{{3D}}$ to 38.36\%. This suggests that geometric consistency between the projective transformations significantly aids in accurate 3D object localization. Multi-vehicle Plane Geometry (MGP) emphasizes the geometric relations among multiple vehicles, improving recall to 39.71\% and bringing the overall ${mAP}_{{3D}}$ to 37.58\%. 

\textbf{Effect of Different 2D Object Detectors.}
The performance of off-the-shelf 2D object detection also affects the matching results of WARM-3D. We explore the impact of different 2D object detectors on the performance of 3D object detection within the WARM-3D framework in Table \ref{tab:Ablation-YOLO}. Specifically, we compare the performances with YOLOv8 and YOLOv9 to understand how variations in 2D detection capabilities influence the effectiveness of 3D detection. YOLOv8 achieved an overall ${mAP}_{{3D}}$ of 38.95\% while, YOLOv9 achieved evidently higher overall ${mAP}_{{3D}}$ of 41.55\%. The results suggest that improvements in 2D detection could significantly enhance the 3D detection in the WARM-3D framework. 

\begin{figure}[bht]
    \centering
    \includegraphics[width=0.48\textwidth]{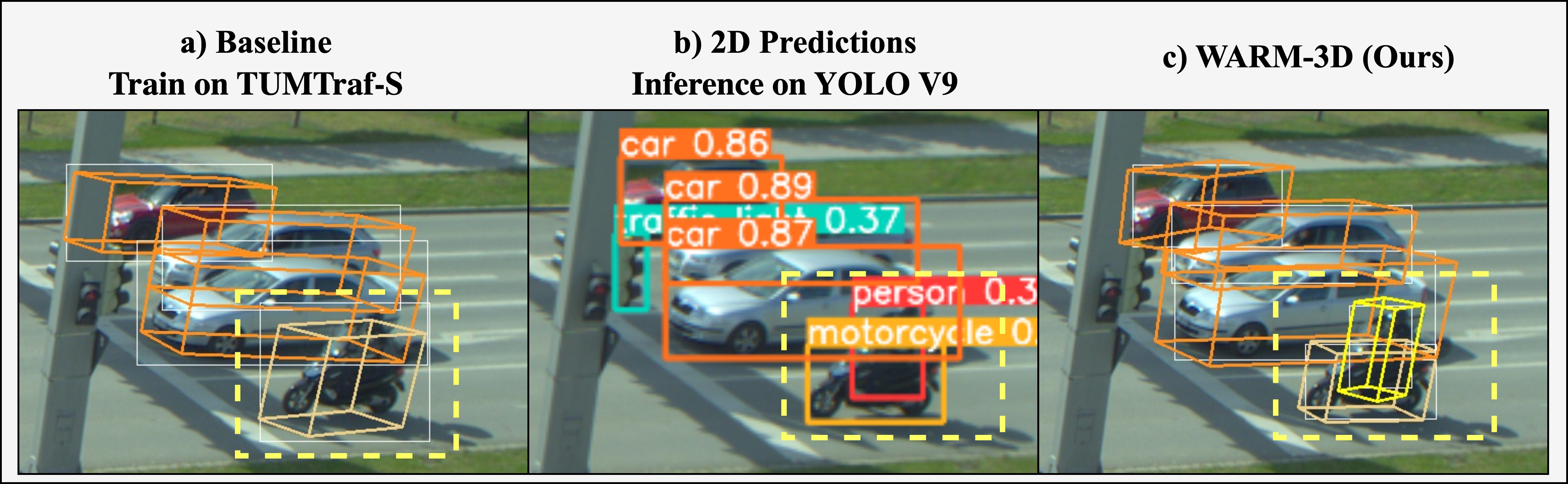}
    \caption{Visualization of the impact of class definition on the WARM-3D performance. As shown in c) column, the WARM-3D results tend to align the class definition of motorcycle and pedestrian to YOLOv9.}
    \label{fig:yolocompare}
    \vspace{-1em}
\end{figure}

\begin{table}[htb!]
\centering
\footnotesize 
\caption{Evaluation result of with different 2D Object Detectors, where YOLOv9 achieves evidently better results.} 
\resizebox{0.48\textwidth}{!}{



    
        \begin{tabular}{cccccccccc}
            \toprule[0.5pt]
            \multirow{2}{*}{\textbf{2D Detectors}} & \multirow{2}{*}{\textbf{Precision}} & \multirow{2}{*}{\textbf{Recall}} & \multicolumn{4}{c}{$\textbf{mAP}_{\textbf{3D}}$}                                                    \\
                                                   &                                     & ~                                & Easy                                             & Mod.           & Hard           & Overall        \\
            \midrule[0.5pt]

            YOLOv8                                 & 39.60                               & \textbf{40.90}                   & 51.78                                            & 34.32          & 23.03          & 38.95          \\

            YOLOv9 \cite{wang2024yolov9}           & \textbf{42.23}                      & 39.84                            & \textbf{55.09}                                   & \textbf{35.17} & \textbf{24.89} & \textbf{41.55} \\

            \bottomrule[0.5pt]
        \end{tabular}
}

\label{tab:Ablation-YOLO}
\vspace{-1em}

\end{table}

Besides, we observe that the domain adaptation performance is also affected by the class definitions, as illustrated in Fig. \ref{fig:yolocompare}. In the definition of TUMTraf-S, a motorcycle includes both the motorcycle and the rider, whereas in YOLOv9, these two are separate classes. After applying WARM-3D, the model tends to align with the class definitions of the 2D object detector.


\begin{figure}[bht]
\vspace{-3mm}

    \centering
    \includegraphics[width=0.48\textwidth]{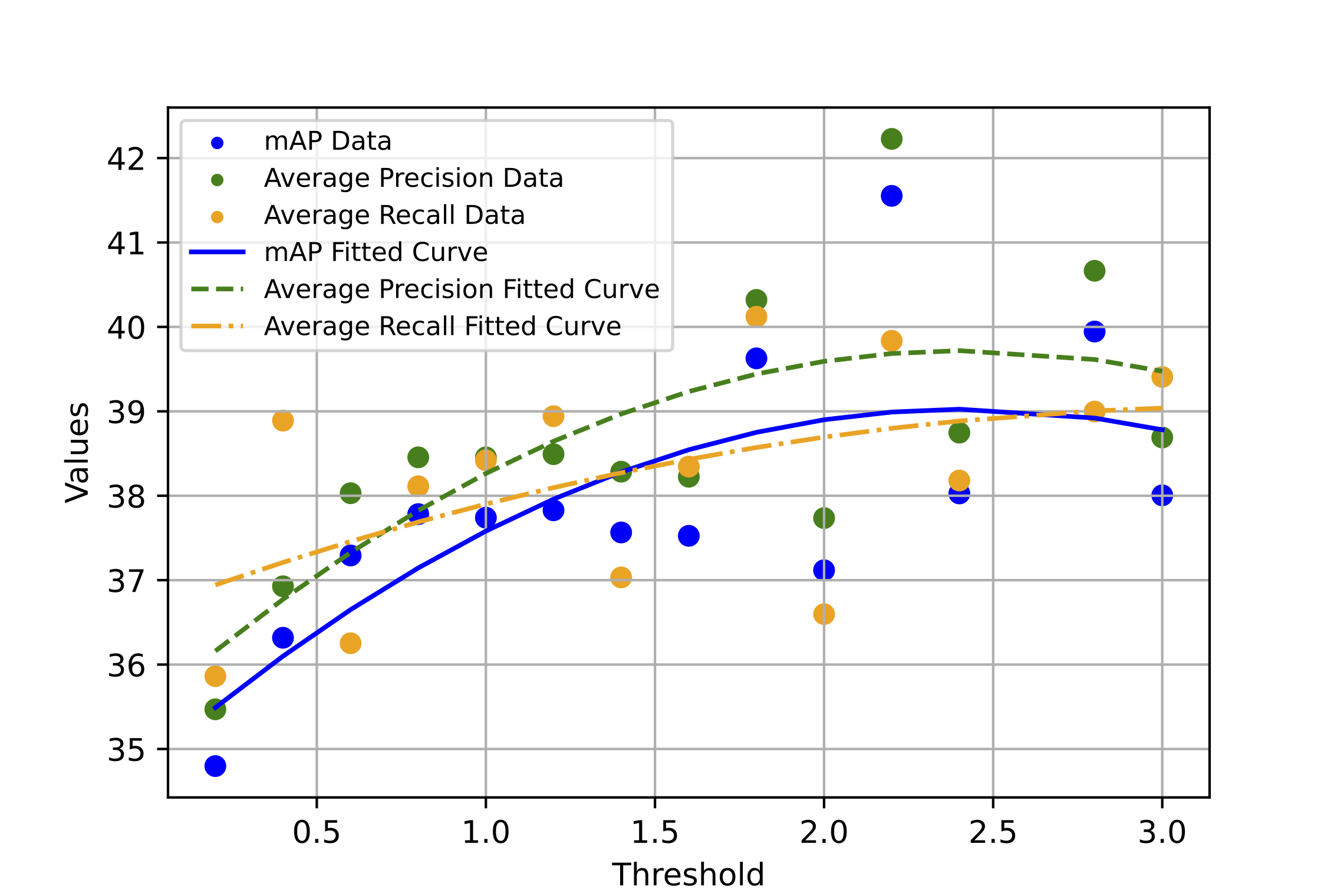}

    \caption{Results of ${mAP}_{{3D}}$, Average Precision, and Average Recall with respect to the Confidence-Aware Bipartite Matching thresholds. The curves indicate the optimal performance observed at moderate threshold values.}
    \label{fig:metrics_vs_threshold}
\end{figure}

\textbf{Threshold Analysis of Confidence-Aware Bipartite Matching.}
In the WARM-3D framework, the CABM module is critical in aligning 2D and 3D predictions by adjusting confidence thresholds. The threshold range is between $[0, 3]$. Based on the results, we set the threshold at 2.2, which yields the optimal overall performance, achieving an ${mAP}_{{3D}}$ of 41.55\%. The result is shown in Figure \ref{fig:metrics_vs_threshold}. 



\section{Conclusion and Future Work}

In this study, we propose WARM-3D, a concise and effective weakly supervised domain adaptation framework for roadside monocular 3D object detection. We present TUMTraf Synthetic Dataset, and show that WARM-3D can achieve excellent Sim2Real domain adaptation performance both quantitatively and qualitatively. We show that WARM-3D can even achieve performance close to the Oracle baseline with only synthetic dataset and 2D real-world weak labels. We also demonstrate that the WARM-3D framework can be easily adapted to unseen new samples, making it possible to train a real-time, stable, and robust monocular 3D detector with large-scale unlabeled data. Notably, the knowledge of 3D scene understanding of WARM-3D is primarily derived from the source label domain. Therefore, constructing an appropriate synthetic dataset is crucial for successful adaptation when extending to other scenarios.

\addtolength{\textheight}{-12cm}   



\section*{ACKNOWLEDGMENT}

This research was supported by the Federal Ministry of Education and Research in Germany (BMBF) within the project ”AUTOtech.agil” (Grant Number 01IS22088U). We thank Robin Brase for the high-quality 3D modeling of the S110 intersection.  


{
    \bibliographystyle{IEEEtran}
    \bibliography{main}
}

\end{document}